# Sequential Clustering-based Facial Feature Extraction Method for Automatic Creation of Facial Models from Orthogonal Views

Alireza Ghahari, Reza Aghaeizadeh Zoroofi
School of Electrical and Computer Engineering
University of Tehran
Tehran, IRAN
a.ghahari@ece.ut.ac.ir, zoroofi@ut.ac.ir

*Abstract*— **Multiview 3D face modeling has attracted increasing attention recently and has become one of the potential avenues in future video systems. We aim to make more reliable and robust automatic feature extraction and natural 3D feature construction from 2D features detected on a pair of frontal and profile view face images. We propose several heuristic algorithms to minimize possible errors introduced by prevalent non-perfect orthogonal condition and non-coherent luminance. In our approach, we first extract the 2D features that are visible to both cameras in both views. Then, we estimate the coordinates of the features in the hidden profile view based on the visible features extracted in the two orthogonal views. Finally, based on the coordinates of the extracted features, we deform a 3D generic model to perform the desired 3D clone modeling. Present study proves the scope of resulted facial models for practical applications like face recognition and facial animation.**

## I. INTRODUCTION

A stereoscopic vision-based system is the assembly of units such as modeling, quantification, segmentation, fusion, interpretation and visualization. Geometric modeling is primarily concerned with method for the representation and manipulation of curves and surfaces. The more impressive unit, visualization, is concerned with the techniques for the transformation of data into pertinent, computer-generated images. The immersive environments, as the baseline, provide engineers and scientists with a means for interacting with massive and complex 3D data, often rendered stereoscopically in a virtual interaction space.

There are many approaches to enable creation of facial models in a virtual world, but two stand out. Firstly, laser scanning is used to register the 3D shape of human head. This approach focuses on reconstructing a remarkable shape, but the biggest drawback is that they provide only the shape without structured information [1]. Secondly, a more convenient way of 3D objects creation is reconstruction from 2D-photo information. Some researchers are using a stereo pair of images or video sequences to create a facial model by finding pixel correspondence between cached frames or stereo-pair. Gökberk et al. [2], as 3D facial features, compared the use of 3D point coordinates, surface normals, curvature-based descriptors, 2D depth images, and facial profile curves. Analysis of decision-level fusion techniques such as fixed-rules and voting schemes is also pinpointed. Others use frontal and profile view to reconstruct a 3D facial model [3,4]. The problem is how to make 3D data from 2D points on a front view and on a side view. However, feature positions at facial landmarks give a lot of information of the human characteristic and features provide the key to build the 3D structure of the persons face. Therefore, the feature extraction on views and how to make correspondence between them have been profound objectives when making a robust facial model. To detect 2D features, some methods have drawbacks such as too few points to guarantee individualized shape from a very different generic head, while others make much use of filtering operators such as Gabor wavelet based filter banks [5].

In this paper we present an efficient method for automatic feature extraction and 3D face modeling in the cases of non-perfect orthogonal condition and non-coherent luminance of frontal and profile view images.

The rest of this paper is organized as follows. In Section 2, we propose a global matching scheme to find locations of features. In Section 3, we describe a facial feature extraction method for detailed matching. In Section 4, the features in the profile views are estimated by utilizing the features obtained from the frontal view. Section 5 shows 3D model deformation results and Section 6 presents some discussions about the efficacy of proposed method in 3D face modeling. Finally, in Section 7, we offer a brief conclusion.

## II. FEATURE GLOBAL MATCHING

### A. Preprocessing

The input images are two views of a person. For each case in the MIT-CBCL database [6], we employ one frontal view and one side view as the inputs. These images are taken under controlled light variations. The preprocessing step comprises of face scale normalization and histogram equalization operations to reduce the probable effects of non-perfect orthogonal and non-coherent luminance conditions. In addition, we assign a rectangular region of interest (ROI) that holds the head in the frontal view in all frames.

### B. Feature Assembly Localization

In this step, we estimate the global position of facial features. We assume the face consists of several structures such as the eyes, lips, nose, etc. Each feature such as an eye can be regarded as a structural assembly of micro features, which are small units of the feature that are being searched. There is also another implicit assumption pointing out that the features cannot appear in arbitrary arrangements. For instance, given the positions of two eyes, it is obvious that the other feature assemblies can only lie within specific area of facial landmarks. We use such a knowledge-based approach to develop a method for micro features assembly localization.

The eyes' texture is different from the skin and has a higher reflection rate than other features. In the discriminative $C_b$ channel of the $YC_bC_r$ color space, the reflection in the pupil may interrupt the processing trend and may strengthen the edges around an eye. Wrinkles on the eyelid may also hinder the localization process. Besides resolving these issues, using the $C_r$ channel makes the method robust to low-quality images.

Wide variety of works accomplished by the template-based matching methods, moderate filtering techniques, statistical Bayesian approaches, and error back-propagation neural networks for feature localization results from the extremely case-dependent image segmentation problem. They are not reliable since their parameters are too sensitive depending on each individual's facial image.

In this work, we use an extended form of the BMCA[1] [8], the sequential cluster detection algorithm (SCDA), adapted for cases in which feature clusters are not properly represented by a single representative.

### C. Sequential Cluster Detection Algorithm (SCDA)

Having utilized basic morphological operators (opening and closing) on the thresholded $C_r$ channel of the frontal view, the feature assembly clusters are determined using one sequential clustering scheme according to following steps:

(a) The V(**x**) neighborhood of one non-processed micro feature, **x**, is determined.

(b) **If** V(**x**) contains at least $\alpha$, a threshold of the density of the V(**x**), micro features then

- One search loop creates a new cluster that includes:
  - The point **x**
  - All points **y** for which there exist a sequence of $c_y$ points, $\mathbf{y}_k$ , k = 1 , ... , $c_y$, such that $\mathbf{y} \in V(\mathbf{y}_1)$, $\mathbf{y}_k \in V(\mathbf{y}_{k+1})$, k= 1 , ... , $c_y - 1$, $\mathbf{y}_{cy} \in V(\mathbf{x})$.
  - The executed points are considered as processed.

**Else**

- **x** is considered as processed.

[1] Binary Morphology Clustering Algorithm

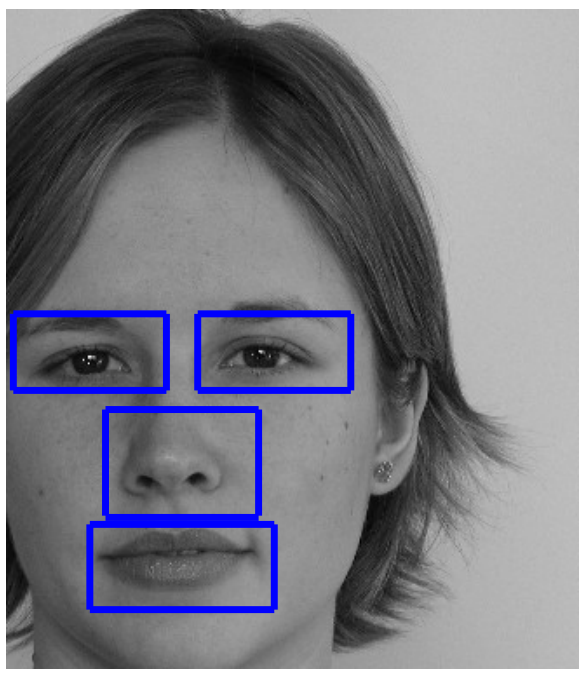

Figure 1. Features assembly sub-images of a sample from the MIT-CBCL database.

Allocating sub-images to feature clusters is accomplished by some parameter definitions related to within scatter matrices of extracted clusters. These parameters are obtained through training on a set of facial patches. In this stage, Left eye window is established. Then, considering some underlying information of $C_b$ channel, the right eye window is localized. Finally, prior knowledge about the face geometry in conjunction with feature clusters scattering details is used to build the nose and mouth windows as shown in Fig. 1.

## III. Feature Extraction Framework in the Frontal View

In the global matching, the facial feature landmarks are extracted automatically. In this section, we intend to automatically extract 60 2D feature points, those which are the most characteristic points used to represent a face, from the both frontal and profile views. Fig. 2a shows 60 marked feature points in the frontal view and Fig. 2b shows 31 corresponding feature points in the visible profile view.

Using an edge detector operator, our purpose is to emphasize the representative weak edges and force it to remove other edge fragments. Although the Canny edge detector is famous for the pure edge extraction and is robust to extract weak edges, it does not perfectly extract the tokens. We provide edge linking for the Canny-processed images as a by-product of one double thresholding scheme to reduce the effect of false positive errors. Fig. 3 shows the processed feature point landmarks for the image in Fig. 1. To extract the 2D feature points picked out from the frontal view, the hierarchical clustering approach in Ref. [7] has been adopted. Final results for the frontal view of a database's subject are given in Fig. 4a. Then, the feature locations can be used to detect the shapes in detail by performing the convex hull operation.

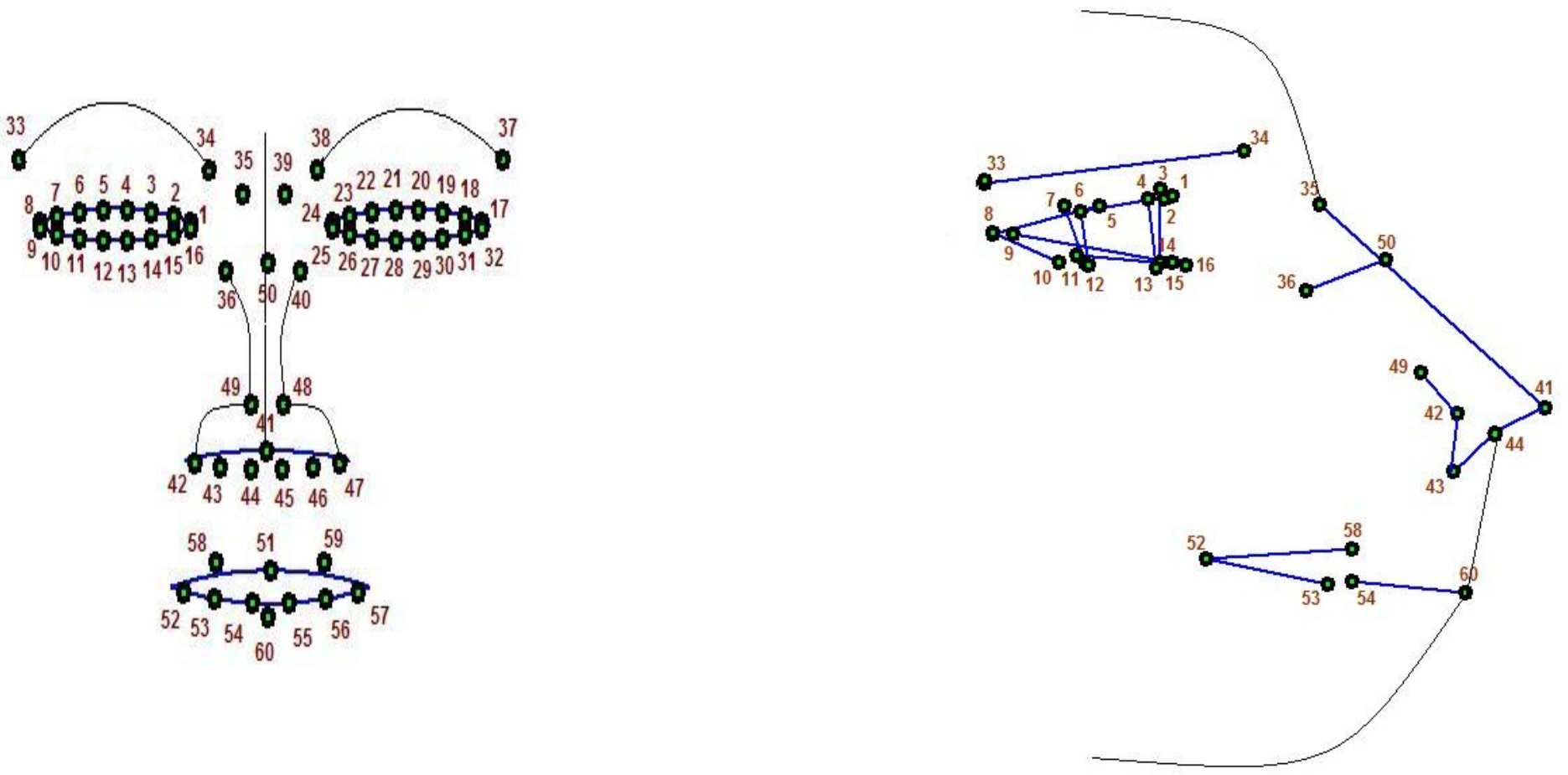

(a) (b)

Figure 2. Marked facial feature points in the frontal and profile view images. (a) Frontal view; (b) Profile view.

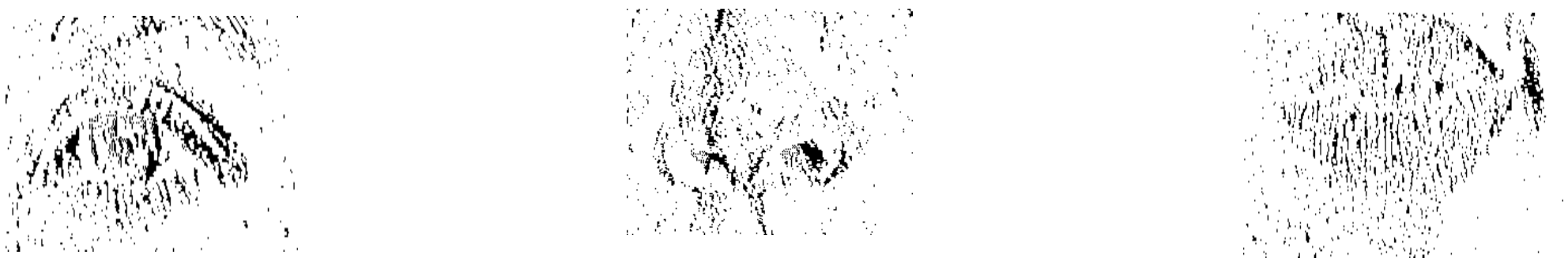

(a) (b) (c)

Figure 3. Facial landmarks processed with modified Canny edge detector. (a) Right eye; (b) Nose; (c) Mouth.

## IV. 3D Computations in the Profile Views

In this section, we deal with the problem of features depth estimation in the profile and hidden profile views. The problem is how to make correspondences between 2D feature points, (X, Y), on frontal view and ones on the visible profile view, (Z, Y). We then make use of symmetry between facial features which scatter at each side of the face and the resulted coordinates from the frontal view and the visible profile view to estimate the Z coordinate of features in the hidden profile view.

### *A. Visible Profile View*

Assuming some functionality such that the Y coordinates on the frontal and profile views share the same value, the challenging issue is the non-perfect orthogonal feasible condition. In order to minimize the errors that may be introduced by the orthogonality constraint, we utilize an algorithm on detected 2D features aiming to decrease the dependency of orthogonality of face images.

In order to search for the feature points Z coordinate around their corresponding Y coordinate in the profile view, a heuristically obtained algorithm, self organizing intensity correlation (SOIC), has been utilized. In this algorithm, the Y coordinate of feature points visible to both cameras in both views construct the primitive search line. Then, an adaptive scheme considers a soft margin for each search line to minimize the induced square error. Using a 3*3 intensity correlation kernel, the algorithm self organizes the margin‘d’ in order to force the solution Z coordinate to lie toward the middle of the d=0 solution region in hopes of improving generalization of resulting depth estimator. Considering the margin d, the correlation result with minimum value determines the Z coordinate of each feature point in the profile view. Fig. 4b shows the extracted feature points which correspond to the right (visible) side of the Fig. 4a.

### *B. Hidden Profile View*

After fulfilling 3D computations in the visible profile view, we estimate the coordinates of the features hidden in the frontal view based on the 2D facial features extracted in the frontal view and in the visible profile view. Making use of facial symmetry between the visible (r) and hidden (l) sides, with a 3D origin established at the middle distance between two eye centers (o), the 3D distance between the origin to one of the symmetrical features at one side of the face ($d_{or}$), is the same or close to the distance from the origin to the corresponding feature at the other side of the face [3]. Consequently, the relative depth of each feature point in the hidden side of the profile view, is calculated using

$$Z_l = \sqrt{d_{or}^2 - (X_l - X_o)^2 - (Y_l - Y_o)^2} + Z_o. \quad (1)$$

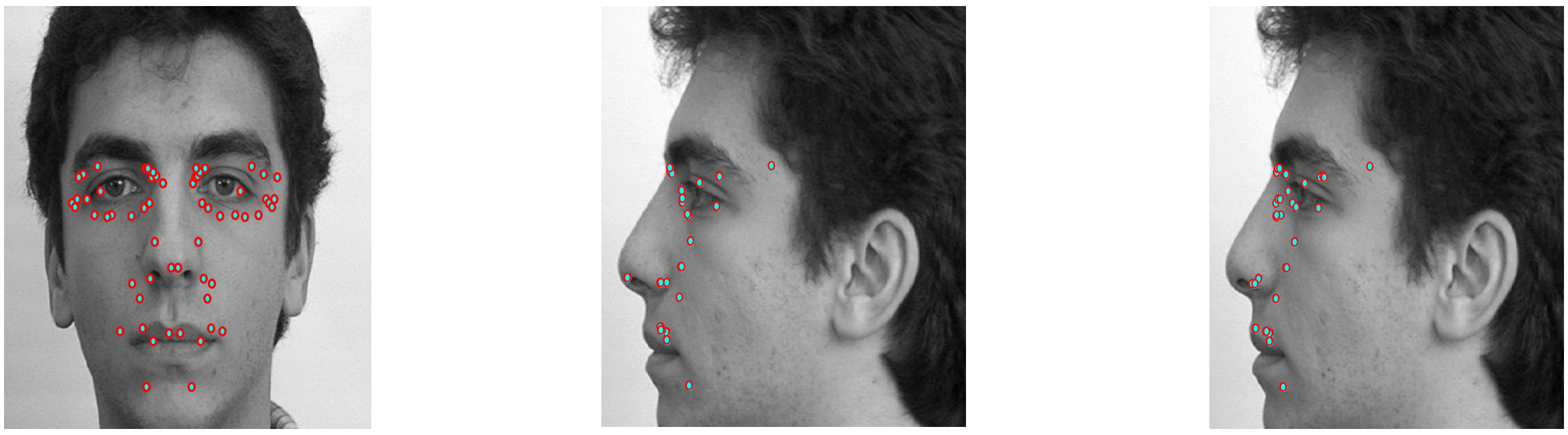

(a) (b) (c)

Figure 4. Final representation of the facial features extracted from frontal and profile view images. (a) Frontal view; (b) Profile view; (c) Hidden profile view.

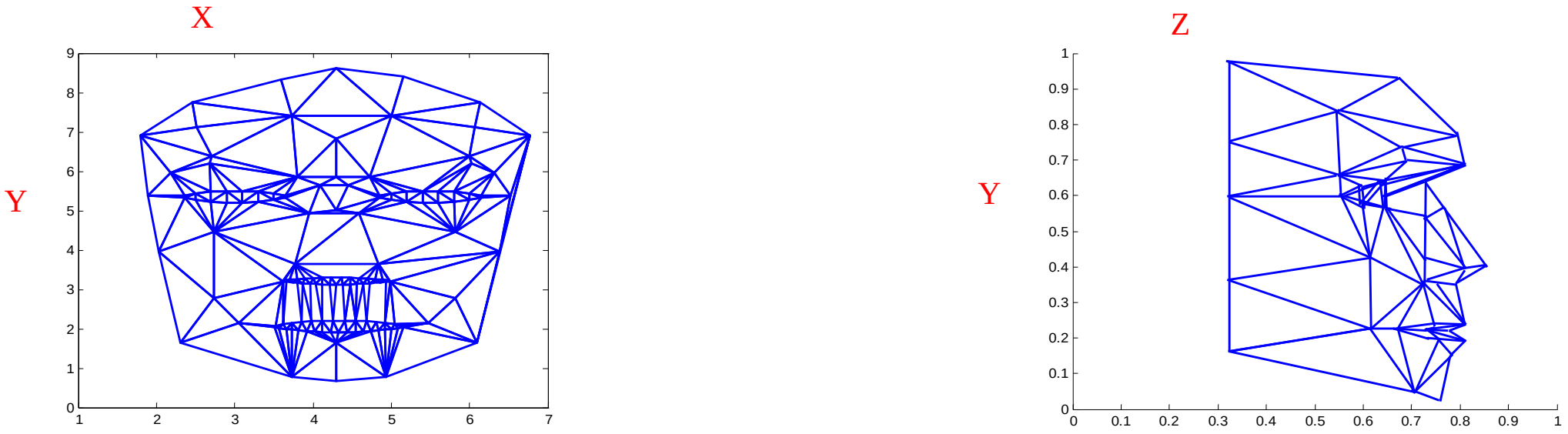

Figure 5. Generic 3D model.

The result gives a predefined number of 3D features created from 2D features detected on two face images of a subject. Fig. 4c shows the computed feature points which correspond to the left (hidden) side of the Fig. 4a.

## V. Generic Model Modification Results

Using 3D feature points as control points for deformation, we deform a 3D generic model to create a 3D cloned model. A host of 3D generic face models are available in the literature. In order to design multifaceted patches, we should specify the coordinates of each unique vertex and a matrix that specifies how to connect these vertices to form the faces. Using NURBS[2] curve segments for vertices definition and Delaunay triangulation solution as the connection principle, a modified version of the well-known Candide model with 140 vertices and 264 surfaces has been designed. Fig. 5 shows the resulted generic face model. Then we modify the shape of the generic model to bring its vertices as close as possible to the corresponding 3D coordinates of the feature points calculated in the section 4. We use the procrustes analysis, summarized in Ref. [3], and the Dirichlet Free-Form Deformation, which belongs to a wider class of geometric deformation tools, as modification functions to probe the results. Before applying the deformation methods, we use translation and scaling transformation to bring obtained 3D feature points to the generic model's 3D space. Fig. 6 shows the results for 2 out of the 6 cases that we considered. The figure shows frontal and profile views followed by the corresponding views of the final virtual clones. We should point out that the deformed 3D model may not resemble a given face subjectively; however the displacement of the features is unique to the given face.

## VI. Discussion

This section embraces a brief discussion about the applications of the current research and some experimental results considering the performance of the proposed automatic 3D face modeling system. 3D object recognition has received plentiful attention in the past few years. The efforts focus on developing representations and algorithms that support the recognition of three-dimensional objects in complex scenes. To create a 3D cloned model, a dual process should be taken into account. The first is to extract discriminating features on the frontal and profile view images. The second is to create a 3D individualized facial model using extracted feature information. We stress the fact that a host of feature points are the best representatives since more information from images lead to better adaptation of an individual. Also, as the reconstructed 3D face inherits the same structure from the generic model with all parametric information, it can be animated immediately with parameters extracted from an automatic facial expression analysis system [7]. At the highest level, animation is controlled by a script containing speech and emotions with their duration. Another application of 3D facial cloning is 3D face recognition. Recent literature surveys of face recognition focus on recognition based on 2D data, 3D data, and 2D+3D data fusion. Training a moderate distance-related classifier, Ansari and A-Mottaleb [3] calculate the distances between each of the 29 feature vertices in the test model and the corresponding vertices in the database models. Testing their algorithm with 26 faces in the database, a recognition rate of about 96.2% achieved. Automatic facial feature extraction plays a vital role in robust 3D face modeling from a pair of frontal and profile face images. On some models, features are not easy to detect even in a semiautomatic way. We incorporate additional subtle feature points into the primitive assembly of feature points to enable the algorithm to handle un-calibrated images. We are not only concern about the mean square error (MSE) of the transformation between the feature points from our database and the corresponding feature vertices of the 3D model, but we also expect that the deformation functions deserve acceptable processing time for modifying a 3D generic model to create a 3D cloned model. There are two experiments conducted in this section. In the first experiment, averaging the results over 10 images form 5 persons of our database [6], the normalized MSE of the transformation, with respect to the number of extracted feature points, has been calculated. The result is shown in Fig. 7 for the specified number of feature points. The extension of Ansari's method has been obtained by using 3D interpolation between the initial 29 feature points. Up to the specified number of feature points (≅ 70), Ansari's method outperforms our method, but ours surpasses his work as the number of feature vertices increases. Also, using the DFFD deformation tool produces slightly better results than procrustes program. The second experiment is constructed to evaluate the processing time allocated to modification process. Fig. 8 presents the processing time resulted from two discussed modification approaches. In the lower bounds of feature point's cardinality (i.e., < 90), the procrustes program is faster than DFFD tool, but the situation reverses in the upper bounds. The adaptation time can be regarded as an outstanding estimation for the total process time of creation of the 3D facial models. Our system is implemented in Matlab on a Pentium IV with 2.20 GHz PC.

[2] Nonuniform, Rational B-Spline

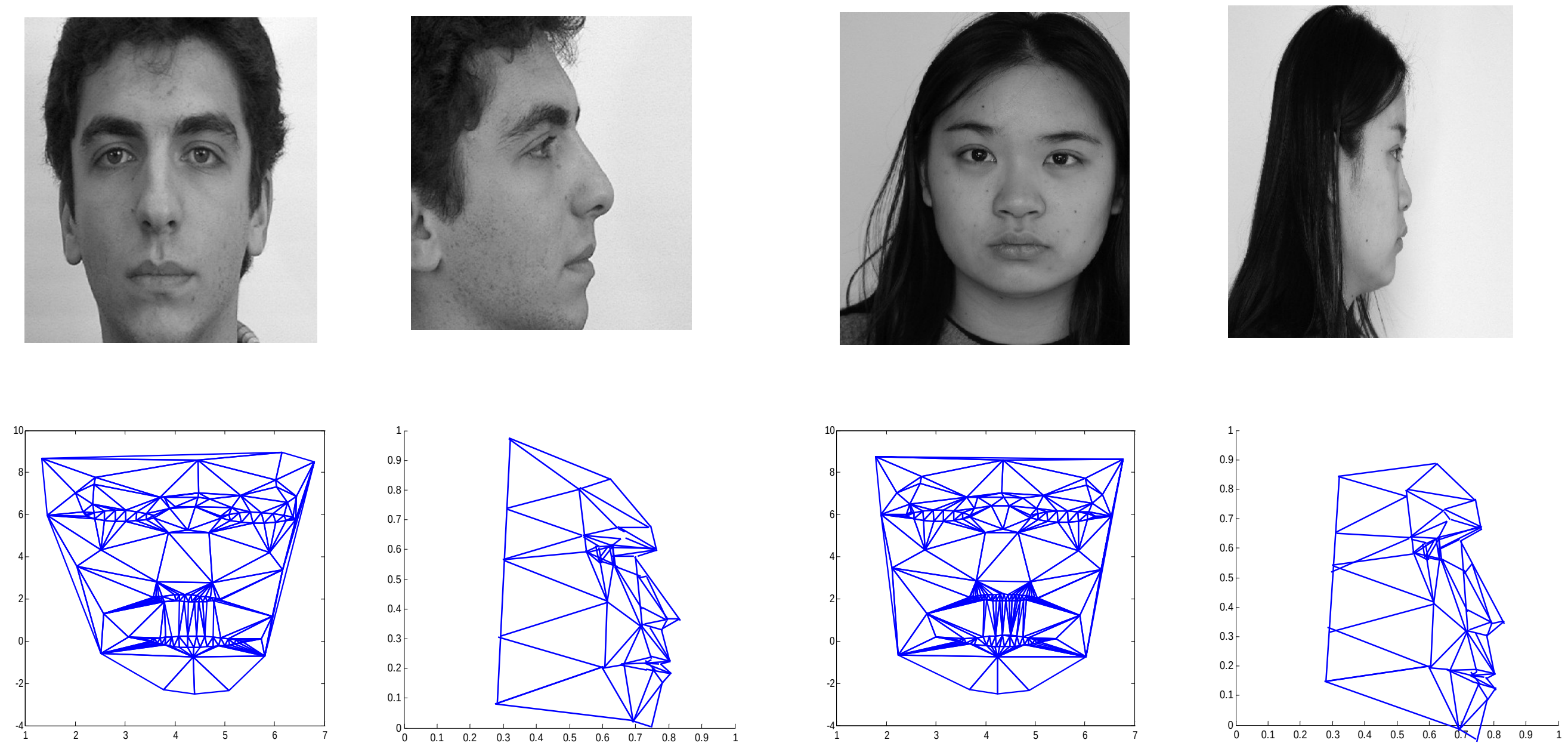

(a) (b)

Figure 6. Input face images and 3D cloned models constructed by procrustes analysis.

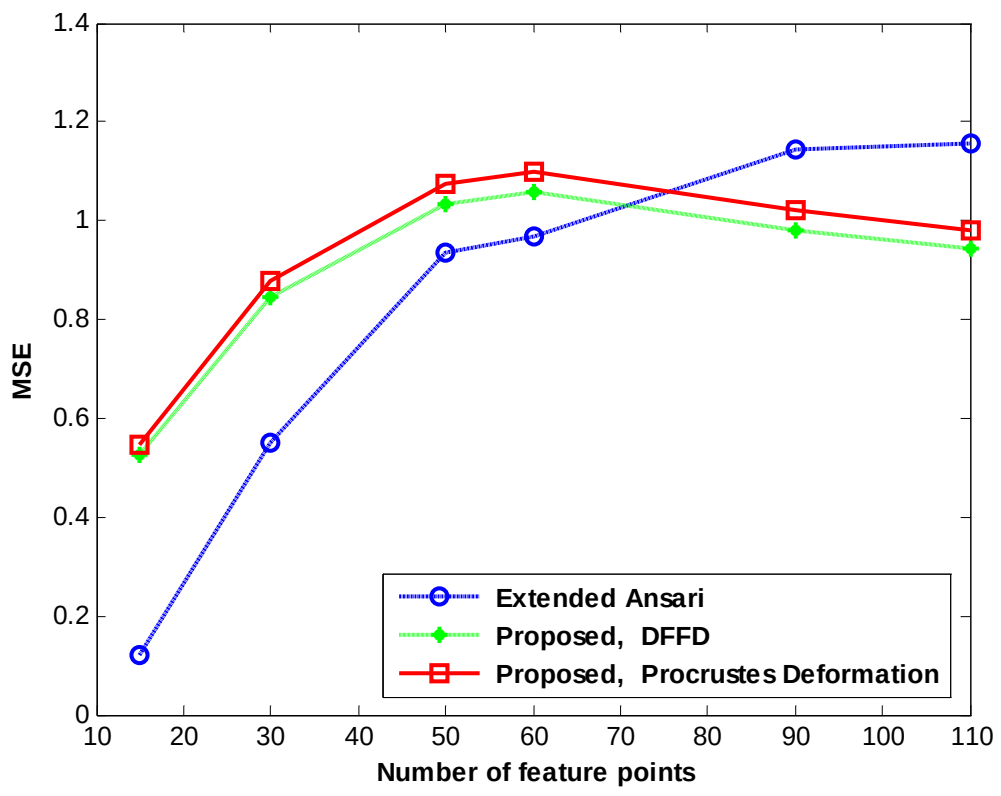

Figure 7. Comparison of the ensemble-averaged MSE performance.

## VII. Conclusions and Future Works

We described an automatic framework for facial feature extraction and robust 3D face modeling. We tried to modify the shape of an already well-prepared model to be individualized using feature information extracted from 2D features of frontal and profile view images. Firstly, we presented the sequential cluster detection algorithm (SCDA) thanks to the feature global matching and feature points localization. After this process, employing the principle of one-to-one correspondence, the 3D coordinates of facial feature points visible in both face images were extracted using the proposed self organizing intensity correlation (SOIC) algorithm. Finally, 3D computations of the features hidden in the profile view enabled us to deform the corresponding facial vertices of the generic 3D model. Some viable considerations were determined to evaluate the overall performance of the facial model creation. The result was remarkable enough to explore real-life applications of 3D face modeling in fields like face recognition and facial animation. Our future study mainstreams are modification of the present approach to build the database of 3D models for more subjects and considering an animation structure to make the cloned models animated in a virtual world.

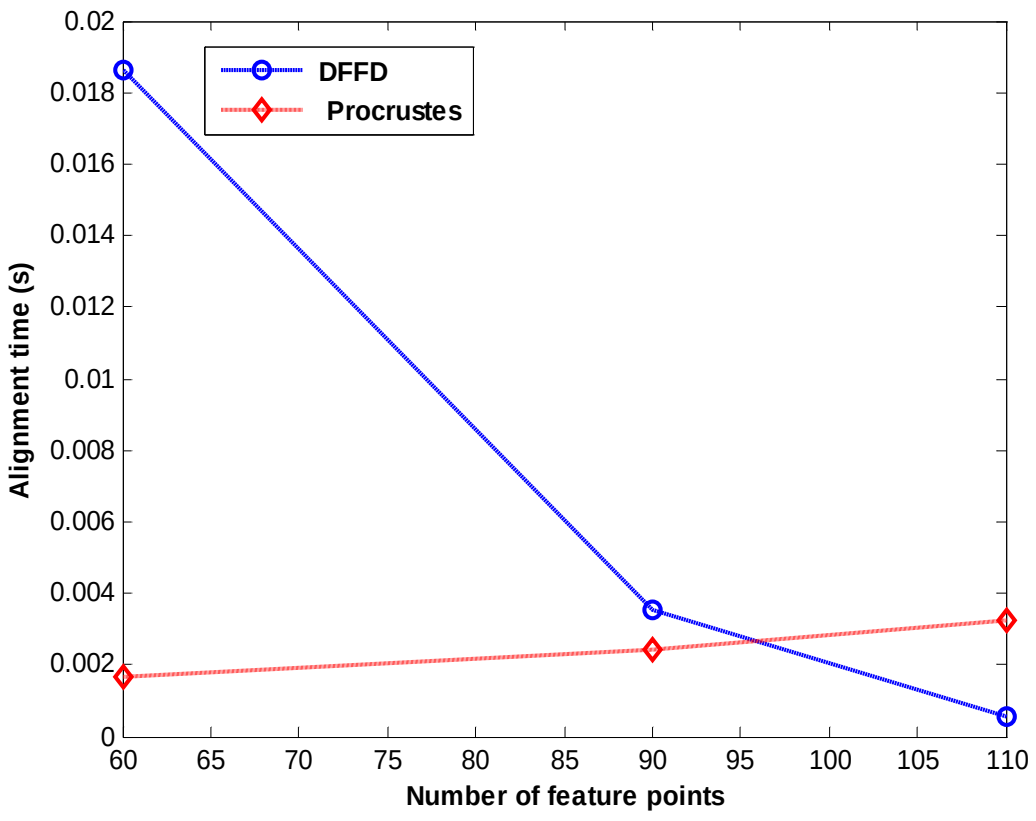

Figure 8. Comparison of the 3D generic model adaptation time.

## Authors Profile

Alireza Ghahari received the Bachlor of Science degree in electrical engineering from the University of Sharif, Tehran, Iran, in 2007. He is currently pursuing his Master of Science degree in electrical and computer engineering at University of Tehran, Tehran, Iran. His research interests are multi-dimensional signal processing and pattern recognition.

Reza Aghaeizadhe Zoroofi received his Ph.D. in medical science from the University of Osaka, Osaka, Japan, in 1998. He is an associate professor in the department of Electrical and Computer Engineering, University of Tehran, where his research is concerned with image processing, advanced computer graphics, and medical software development. He has been an associate editor of several conference proceedings from 2003.